\documentclass{article}
\usepackage[preprint]{neurips_2023}

\usepackage{graphicx}
\usepackage{amsmath}
\usepackage[utf8]{inputenc} 
\usepackage[T1]{fontenc}    
\usepackage{hyperref}       
\usepackage{url}            
\usepackage{booktabs}       
\usepackage{amsfonts}       
\usepackage{nicefrac}       
\usepackage{microtype}      
\usepackage{xcolor}         
\usepackage{cleveref}
\usepackage{subcaption}

\setcitestyle{numbers,square}
\newcommand{\x}{{\boldsymbol{x}}}
\newcommand{\y}{{\boldsymbol{y}}}
\newcommand{\bin}{{\text{Bin}}}
\title{Learning Robust Precipitation Forecaster by Temporal Frame Interpolation}

\author{
   Lu Han\thanks{Equal contribution}, Xu-Yang Chen\footnotemark[1], Han-Jia Ye, De-Chuan Zhan\thanks{Corresponding author}\\
   National Key Laboratory for Novel Software Technology, Nanjing University, China\\
   School of Artificial Intelligence, Nanjing University, China \\
   \texttt{\{hanlu, chenxy, yehj\}@lamda.nju.edu.cn, zhandc@nju.edu.cn}
}

\begin{document}

\maketitle

\begin{abstract}
Recent advances in deep learning have significantly elevated weather prediction models. 
However, these models often falter in real-world scenarios due to their sensitivity to spatial-temporal shifts. 
This issue is particularly acute in weather forecasting, where models are prone to overfit to local and temporal variations, especially when tasked with fine-grained predictions.
In this paper, we address these challenges by developing a robust precipitation forecasting model that demonstrates resilience against such spatial-temporal discrepancies. 
We introduce Temporal Frame Interpolation (TFI), a novel technique that enhances the training dataset by generating synthetic samples through interpolating adjacent frames from satellite imagery and ground radar data, thus improving the model's robustness against frame noise.
Moreover, we incorporate a unique Multi-Level Dice (ML-Dice) loss function, leveraging the ordinal nature of rainfall intensities to improve the model's performance. 
Our approach has led to significant improvements in forecasting precision, culminating in our model securing \textit{1st place} in the transfer learning leaderboard of the \textit{Weather4cast'23} competition. 
This achievement not only underscores the effectiveness of our methodologies but also establishes a new standard for deep learning applications in weather forecasting.
Our code and weights have been public on \url{https://github.com/Secilia-Cxy/UNetTFI}.
\end{abstract}
\section{Introduction}

Weather forecasting has long been a critical domain with far-reaching implications across various sectors, including agriculture, transportation, and public safety. Accurate weather predictions are essential for planning and decision-making, helping to mitigate the impacts of adverse weather conditions and capitalize on favorable ones.
Traditionally, weather forecasting has relied on a combination of meteorological science and empirical techniques. These methods often include numerical weather prediction (NWP) models~\cite{lynch2008origins} that simulate the atmosphere's physical processes. While effective, these approaches can be limited by computational constraints and the inherent complexity of weather systems.

The advent of machine learning (ML) has ushered in a transformative era in weather forecasting. ML techniques, leveraging large datasets and powerful computational resources, have shown remarkable potential in improving the accuracy and efficiency of weather predictions. These advancements are particularly significant in the context of complex, dynamic systems like weather, where traditional models struggle to capture the intricate, non-linear interactions within the atmosphere.
For example, the use of ConvLSTM to capture spatiotemporal correlations has shown promising results in weather forecasting, garnering significant attention~\cite{shi2015convolutional}. Innovations such as MetNet~\cite{sonderby2020metnet}, which processes inputs from an area larger than the target region, have set a new standard in deep learning-based weather forecasting. A notable limitation is that, unlike human vision systems~\cite{hendrycks2019benchmarking}, the performance of data-driven deep learning models significantly deteriorates when faced with data differing from the training phase distribution. Weather forecast models are particularly sensitive to spatial-temporal shifts due to varying regional weather and climate characteristics, making practical application challenging.

 The \textit{Weather4cast} competition~\cite{gruca2022weather4cast} focuses on the solving of spatial-temporal shifts in precipitation forecasting tasks. It has advanced modern algorithms in AI and machine learning through a highly topical interdisciplinary competition challenge: The prediction of hi-res rain radar movies from multi-band satellite sensors, requiring data fusion of complementary signal sources, multi-channel video frame prediction, as well as super-resolution techniques. In 2023, this year's competition builds upon its predecessor's challenge, introducing a novel aspect: the prediction of actual rainfall amounts, a departure from the previous focus on binary rainfall masks, posing a more fine-grained task and requiring more accurate and robust forecasting models when faced with spatial-temporal shifts~\footnote{\url{https://weather4cast.net/}}.  
 
 The transfer learning benchmark for the \textit{Weather4cast} competition involves utilizing training data collected from multiple regions, specifically R15, R34, R76, R4, R5, R6, and R7, spanning the years 2019 and 2020. The performance of the model will be evaluated on unseen data from regions R8, R9, and R10, during the years 2019 and 2020, as well as on data from all ten regions in the year 2021.
 The shift in regions and years requires the model to learn the meta-knowledge that is shared among all the regions and years. The model should also be robust against the variation in the data. 
 
 While traditional weather forecasting methods have laid a foundational baseline, they are limited in managing the complexity and noise in satellite and radar data. Our model, consisting of a novel Temporal Frame Interpolation (TFI) method, a multi-level dice loss, and carefully chosen training strategies, provides a novel solution for these limitations. TFI enhances forecast robustness by interpolating between adjacent frames of satellite and ground radar data, creating synthetic samples. Such mixing of training samples overcomes the insufficient samples caused by the sample rate, as well as increasing the robustness of models against unseen variations on images. The newly proposed multi-level dice loss extracts the ordinal relationship between different rainfall rates, enabling more accurate prediction. Our method has achieved significant success, securing first place in the transfer learning leaderboard of the competition. 

Our main contributions are as follows:

\begin{itemize} 
\item We propose a novel method called Temporal Frame Interpolation (TFI), which enhances the robustness and accuracy of precipitation forecasting models by creating synthetic samples between adjacent frames. 

\item We propose Multi-Level Dice (ML-Dice) loss that takes the ordinal relationship between rainfall levels into account, achieving better forecasting results.

\item We explore multiple combinations of training strategies, including input cropping, output cropping, and geometric augmentations. By integrating these strategies along with our proposed TFI and multi-level dice loss, our model can achieve the best result on the transfer learning leaderboard of \textit{Weather4cast'23} competition, showing the practical applicability and superiority of our method over existing approaches under rigorous and competitive conditions. 
\end{itemize}

\section{Problem Statement}
In the NeurIPS 2023 \textit{Weather4cast} competition, participants face the intricate task of forecasting rainfall rates. This requires interpreting the temporal progression of multi-band satellite observations, encompassing visible, near-infrared (Near-IR), water vapor, and infrared (IR) spectrums. These multi-spectral satellite measurements originate from the MeteoSat satellites, operated by the European Organization for the Exploitation of Meteorological Satellites (EUMETSAT). The objective is to accurately estimate rainfall rates, which are derived from ground-based radar reflectivity data collected by the Operational Program for Exchange of Weather Radar Information (OPERA) radar network.

Formally, the precipitation model is given a sequence of multi-band satellite image $X_t = (\x_{t}, \x_{t+1}, \x_{t+2}, \x_{t+3})$, where each $\x_t \in \mathbb{R}^{11 \times 252 \times 252}$ is the 11-channel satellite image with 252×252 pixels at time $t$. 
Each satellite image captures a 15-minute interval. The OPERA ground-radar rainfall rates as targets $Y_t = (\y_{t+4}, \y_{t+5} ,\dots, \y_{t + 3 + T})$ with each radar image $\y_t \in \mathbb{R}^{252 \times 252 }$ are of T-time steps ($T=32$ for core tack and $T=16$ for transfer learning track). The radar images are also 252×252 pixel patches but note that the spatial resolution of the satellite images (12 km) is about six times lower than the resolution of the ground-radar rain rate products (2 km). The ground-radar rainfall products with a 252×252 pixel patch correspond to a 42×42 pixel patch-centered region in the coarser satellite resolution. A large input area surrounding the target region can supply sufficient synoptic-scale and mesoscale context information for a prediction of future weather, particularly for long lead times. The precipitation forecasting model $f$ needs to predict the future rainfall rates $Y$ based on the short history given by multi-band satellite image $X$, i.e., the model $f: \mathbb{R}^{4 \times 11 \times 252 \times 252} \mapsto \mathbb{R}^{T \times 252 \times 252}$.

\section{Method}

In this section, we first give an introduction to the proposed Temporal Frame Interpolation (TFI) method in~\cref{sec:tfi}.
Then, we describe the Multi-Level Dice loss (ML-Dice) for exploiting the ordinal relationship between target rainfall rates in~\cref{sec:loss}.
Next, we thoroughly describe the other details of our method, including the selection of network structure, input/output processing, and augmentation strategy in~\cref{sec:model}.

\subsection{Temporal Frame Interpolation}
\label{sec:tfi}
\begin{figure}
    \centering
    \includegraphics[width=\linewidth]{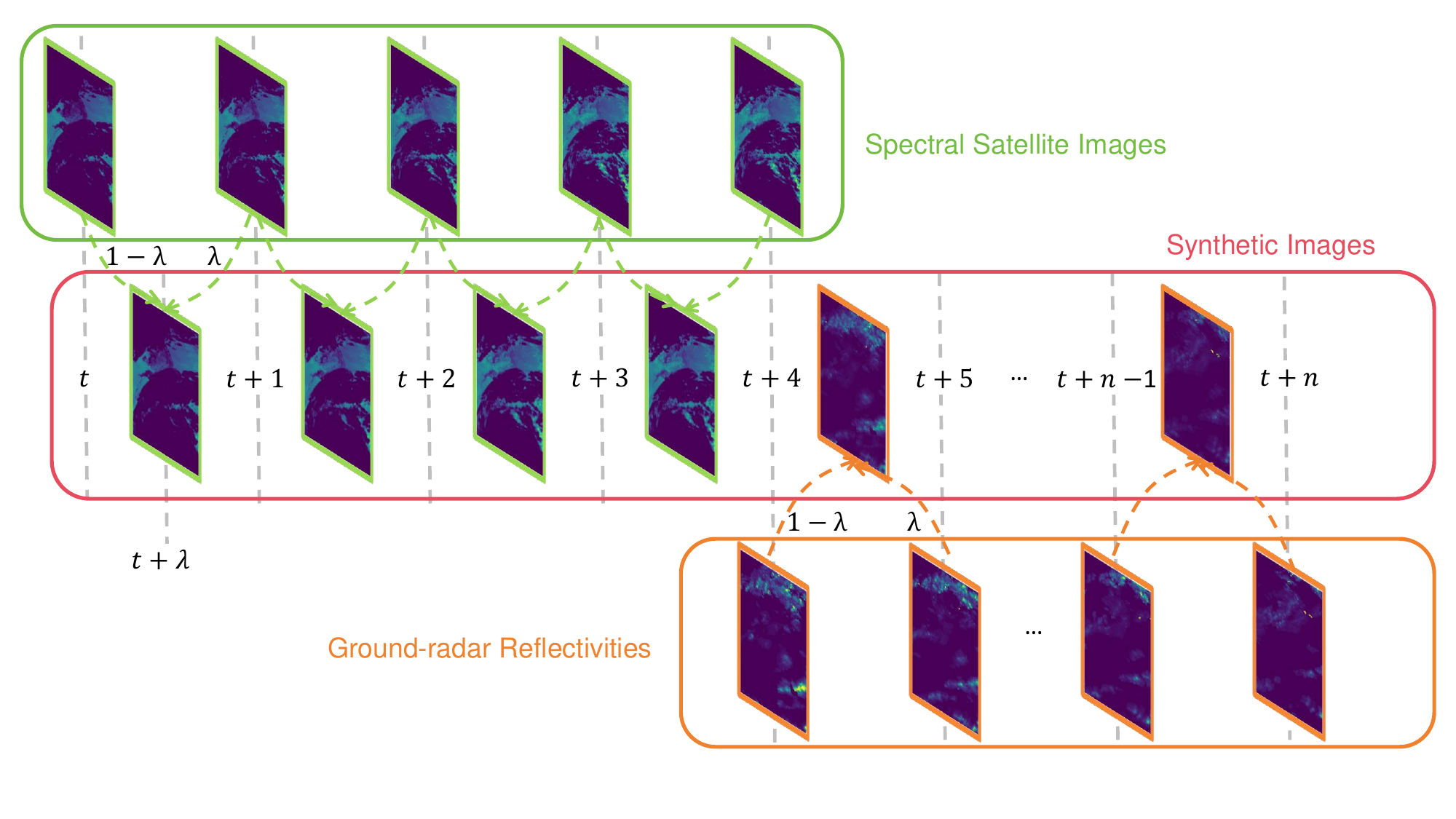}
    \caption{Illustration of Temporal Frame Interpolation (TFI). TFI overcomes the insufficient samples caused by the finite sampling rate. It creates new synthetic samples within the sampling interval by interpolating adjacent frames of both input satellite images and output ground radar data with mixup factor $\lambda$. TFI helps the model better understand the sequence of satellite images and reduces the likelihood of overfitting.}
    \label{fig:tfi}
\end{figure}

In the task of rainfall prediction, the model is tasked with extracting features from high-dimensional sequential image data to predict future high-dimensional rainfall images. Training on such high-dimensional data often poses the risk of overfitting. This is exacerbated by the sampling intervals of satellite imagery; sampling data every 15 minutes leads to insufficient training data for a comprehensive understanding of satellite image sequences, making it challenging to develop models with robust generalization capabilities.

Existing research indicates that deep learning networks, such as Convolutional Neural Networks (CNNs), often struggle to make accurate predictions on interpolated data, even when it is strikingly similar to the original data before interpolation. This limitation is particularly evident in image processing tasks, suggesting a gap in these models' ability to handle sequentially correlated data effectively.
Traditional mixup methods~\cite{mixup} are effective in addressing overfitting by training models on convex combinations of pairs of examples and their labels. However, mixup techniques are primarily designed for classification tasks and may not be directly applicable to the unique requirements of rainfall prediction. The physical characteristics of different regions and years can vary significantly, making the mixup of different samples potentially misleading. 

To address these challenges, we propose the Temporal Frame Interpolation (TFI) method. TFI operates by performing linear interpolation between two adjacent frames within the same region. Given the spatial consistency and temporal proximity, we posit that the distribution of satellite imagery data does not significantly change between these frames. Consequently, the corresponding rainfall amounts can be assumed to have a linear relationship with the interpolated satellite images. This interpolation not only enriches the training dataset but also reduces the likelihood of overfitting by providing a more continuous representation of temporal changes.

We concrete the implementation of TFI here. During the training time, we extend the sample of satellite and radar images to an additional time step. That is, we additionally add the $\x_{t+4}$ and $\y_{t+3 + T}$ as the additional frame of each sample. Then, like the process of mixup, we perform linear interpolation on both the input and output images with a time step shift:
\begin{equation}
    \hat{X}_{t+\lambda} = (1-\lambda)  X_t + \lambda X_{t+1},\quad
    \hat{Y}_{t+\lambda} = (1-\lambda) Y_t + \lambda Y_{t+1}
\end{equation}
where $\lambda\sim Be(a,b)$ ($Be$ stands for Beta distribution) is the random sampled mixup factor. In our solution, we fix $a=1,b=1$, which is equal to a sample from a uniform distribution ranging in $[0,1]$. 

The overview of the Temporal Frame Interpolation method can be found in \cref{fig:tfi}.

\subsection{Multi-Level Dice Loss} 
\label{sec:loss}
This year's \textit{Weather4cast} competition moves from binary classification to zero-inflated regression. Instead of the conventional RMSE (Root Mean Square Error) typically associated with regression tasks, the competition employs the mean CSI (Critical Success Index) score at various rainfall intensity thresholds, specifically at 0.2, 1, 5, 10, and 15, to reward correct prediction over the whole gamut of rainfall intensities. This switch makes the task more like a multi-class classification task. Although we can still use the regression model to accurately predict the precipitation, existing research has revealed that the regression objective and deep learning models are not robust against noise~\cite{ordinal_regression}, especially for the long-term time series forecasting tasks~\cite{han2023capacity}. Hence, our decision is to formulate precipitation forecasting as a multi-class classification task since the absolute value within each classification bin does not impact the CSI score; rather, the critical factor is the accuracy of classifying data points into their respective bins. Formally, instead of directly predicting the continuous rainfall rate, the model outputs the probability of rainfall belonging to different bins where each bin represents a range of rainfall rates. In this competition, since the thresholds 0.2, 1, 5, 10, and 15 divide the rates into 6 bins, i.e.,  $ \bin =  \left([0,0.2), [0.2,1),[1,5), [5,10),[10, 15), [15,+\infty) \right)$, the model will output a 6-dimensional probabilistic vector for each pixel. Denote the encoder as $g$, then:
\begin{equation}
    P_t= g(X_t) 
\end{equation}
where $P_t\in \mathbb{R}^{T \times 252 \times 252 \times 6}$ represents the probability of belonging to each bin of each pixel. For training the model, we choose the dice loss to alleviate the imbalance problem that lies in the prediction task. However, the dice loss can not fully exploit the characteristics of the fine-grain precipitation task. Thus we next propose a new multi-level dice loss.

Dice loss is a performance metric commonly used in the field of image segmentation, particularly in medical image analysis. It's derived from the Dice coefficient, which is a statistical tool used to measure the similarity between two sets. Dice loss was originally proposed to solve binary classification tasks, a direct way to extend the binary dice loss to the multi-class scenario is to compute the average dice loss of each class like in segmentation tasks~\footnote{We do not weight each class like~\cite{gdl} since the evaluation metric of~\textit{Weather4cast} is the unweighted mean of each class's CSI.}. This kind of Dice loss is computed as follows:
\begin{equation}
    L_{\text{Dice}} = 1- \frac{1}{6} \sum_i^6 \frac{ \sum_n \mathbb{I}(y_n \in \bin_i) p_{n,i}}{\sum_n \mathbb{I}(y_n \in \bin_i) + \sum_n p_{n,i}},
    \label{eq:dice}
\end{equation}
where $p_{n,i}$ s the probability of pixel $n$ belonging to $\bin_i$. However, this Dice loss does not take the ordinal relationship of rainfall rates into consideration. For example, if the model mistakenly predicts an [10, 15) rainfall rate while its true ground label is at [1, 5). This Dice will only punish the output on [1, 5), [10, 15), which is not reasonable and not robust, since the severity of overestimating and underestimating precipitation forecasts is not the same. Overestimating precipitation will not affect the CSI on lower thresholds, but underestimation will affect all the higher thresholds. To overcome the drawbacks, we propose Multi-level Dice loss (ML-Dice). Instead of requiring the model to predict the exact bin of the rainfall, ML-Dice punishes the prediction that does not locate itself right to each threshold. Denoting the vector of thresholds as $\boldsymbol{s} = (0.2,1,5,10,15)$, ML-Dice is computed as:

\begin{equation}
    L_{\text{ML-Dice}} = 1- \frac{1}{5} \sum_i^5 \frac{ \sum_n \mathbb{I}(y_n > \boldsymbol{s}_i) \hat{p}_n(y>\boldsymbol{s}_i)}{\sum_n \mathbb{I}(y_n > \boldsymbol{s}_i) + \sum_n \hat{p}_n(y>\boldsymbol{s}_i)},
\end{equation}
where $\hat{p}_n(y>\boldsymbol{s}_i) = \sum_{m=i+1}^6 p_{n,m}$ is the model's prediction of the probability that the output rate exceeds the threshold $\boldsymbol{s}_i$. 
Unlike the vanilla Dice loss in~\cref{eq:dice}, ML-Dice gradually reduces the loss as the prediction approaches the ground target. For example, the loss reduces if the prediction is corrected from [10, 15) to [5, 10) while the target is [1, 5), since it is additionally correct on threshold 10. Therefore ML-Dice exploits the ordinal relationship between the bins and enables to model to 
better trade-off between suboptimal predictions. The vanilla Dice loss in~\cref{eq:dice} can not exploit such relationships since  
it equally punishes [5, 10) and [10, 15).
To further enhance the robustness of Dice loss, we utilize the $logcosh$ transformation on $L_{\text{ML-Dice}}$~\cite{logcosh}.

\paragraph{Inference.} During inference time, we select the bin with the highest probability and generate the rainfall rates as the median of the selected bin.

\subsection{Other Model details}
\label{sec:model}
\paragraph{Network structure.} 
In the initial phase of the competition, we observed that the 3D U-Net architecture~\cite{3dunet}, as utilized in the official baseline~\footnote{\url{https://github.com/agruca-polsl/weather4cast-2023}}, is strong enough to achieve robust results. However, as the competition progresses, we found that the 2D U-Net~\cite{unet} exhibits swifter convergence and delivers a little superior results. Consequently, we decided to transition to the 2D U-Net architecture.

\paragraph{Input and output processing.} Following last year's winners, we crop the input image to be \textbf{126x126} to avoid redundant information provided by the surrounding region~\cite{2ndwinner}. We also constrain the model to output a small \textbf{42x42} patch and use and simple upsampling strategy to restore the 252x252 prediction~\cite{1stwinner,sonderby2020metnet}.
\paragraph{Augmentation strategies.}Inspired by last year's winning solution on Transfer leaderboard~\cite{seo2022domain}, we have employed geometric transformations for data augmentation. The augmentation methods we selected include \textit{Vertical flip}, \textit{Horizontal flip}, \textit{Vertical flip+Horizontal flip}. These geometric transformations do not change the center of a satellite image and the corresponding target radar image can be achieved by simple transformations without loss of any information. Note that due to the restriction of time, we haven't tested all the possible augmentations. But the chosen have been proven effective.

\section{Experiments}
We conducted all experiments using the training set, core validation set, and held-out set of the \textit{Weather4cast’23 Transfer} dataset.
The performance evaluation metric of all experiments was selected as the mean Critical Success Index (mCSI) over five thresholds 0.2, 1, 5, 10, 15.

\subsection{Experimental Setting}
The datasets used in Weather4cast’23 do not change compared to Weather4cast’22. 
\paragraph{\textit{Training set.}}  The training set encompasses both satellite and terrestrial radar data, encompassing a span of two years, 2019 and 2020, across seven distinct European regions.
The satellite data was sourced from a geostationary meteorological satellite managed by EUMETSAT, while ground truth radar data was acquired from OPERA, a collaborative weather radar information exchange program.
This satellite dataset includes a variety of spectral bands, capturing information across the visible spectrum, infrared, and water vapor channels, such as IR016 and VIS006, among others.
Each spectral frequency provides unique insights due to their distinct interactions with the atmospheric constituents.
For example, the visible (VIS) bands harness solar radiation to yield data on albedo and surface texture but are limited to daylight hours for data collection.
In contrast, infrared (IR) bands detect radiation re-emitted from objects, which can be translated into measures of brightness temperature, while water vapor (WV) bands quantify the water vapor present in the higher atmospheric layers.
These bands are instrumental for differentiating clouds from the surroundings and examining cloud dynamics via single or combined channel analysis.
Satellite imagery has a spatial footprint of 12 km x 12 km, captured at quarter-hourly intervals, whereas radar imagery, with a resolution of 2 km x 2 km, focuses on a central subset of the satellite's coverage.

\paragraph{Transfer test set.} The test set for the transfer leaderboard consists of regions not present in the training dataset (indicating a spatial shift), labeled R08, R09, and R10, in addition to a set of regions from the year 2021 (indicating a temporal shift), including R15, R34, R76, R04, R05, R06, R07, R08, R09, and R10.
\paragraph{Core validation set.} The core validation set consists of data from all the seen regions (R15, R34, R76, R4, R5, R6, and R7) in 2019 and 2020. 
\paragraph{Details on Implementation} As referenced in~\cref{sec:model}, our early competition strategy involved utilizing a variant of the 3D-UNet architecture~\cite{3dunet}, which later transitioned to a 2D-UNet model. It was this latter model that yielded the leading results on the leaderboard. However, certain ablation studies were conducted using the 3D-UNet configuration.
We configured our single GPU batch size at eight and ran the models for a total of 90 epochs.
The initial learning rate was established at 0.0001, with a weight decay set at 0.02 and a dropout rate maintained at 0.4.
The optimization was managed via AdamW, featuring a learning rate scaled down by a factor of 0.9 whenever the validation loss exceeded that of the previous epoch. Furthermore, an early stopping mechanism was enforced to halt training if the validation loss plateaued.
All experiments were executed on 4 Nvidia RTX 4090 GPUs.

\subsection{Main Results}
\begin{table}[htp]
\centering
\caption{Results on \textit{Weather4cast’23 Transfer Learning} leaderboard.}
\begin{tabular}{@{}cccccccc@{}}
\toprule
Team & SandD (ours)    & ALI\_BDIL  & enrflo     & imiracil   & rainai     & sandeepc  &Baseline   \\
\midrule 
mCSI     & \textbf{0.10146} & 0.09552 & 0.07072 & 0.05915 & 0.05842 & 0.05754  & 0.04585 \\
\bottomrule
\end{tabular}
\label{tab:main}
\end{table}
\paragraph{Results} ~\Cref{tab:main} is the \textit{Weather4cast’23 Transfer} held-out leaderboard. As shown in the table, our method achieved the highest performance in the leaderboard with 0.10146 mCSI (mean CSI). This result is over 2 times the baseline result and outperforms most teams by a large margin. Our solution is also the only one that surpasses 0.1 mCSI, showing the accuracy and robustness of our precipitation model against other solutions.

\subsection{Ablation Study}
We validate the effectiveness of our methods on the validation set (R15, R34, R76, R4, R5, R6, and R7 in 2019 and 2020) of the core track. Here lists the mean CSI and mean F1 on the core validation set. The components of our methods include Input Cropping (IC) to 126x126, Output Cropping (OC) to 42x42, and geometric Augmentation (Aug), as well as newly proposed Multi-Level Dice loss (ML-Dice) and Temporal Frame Interpolation (TFI). 
\begin{table}
\centering
\caption{Mean CSI and mean F1 on the validation set of core track.}
\begin{tabular}{lll} 
\toprule
Method                 & mCSI   & mF1     \\
\midrule
UNet2D (ML-Dice+IC+OC+Aug+TFI)       & \textbf{0.1069} & \textbf{0.1738}  \\
UNet2D (ML-Dice+IC+OC+Aug)        & 0.1034 & 0.1701  \\
UNet3D (ML-Dice+IC+OC+Aug) & 0.1014 & 0.1657  \\
UNet3D (Dice+IC+OC+Aug) & 0.0986 & 0.1615  \\
UNet3D (Dice+IC+OC)     & 0.0932 & 0.1539  \\
UNet3D (Dice+IC)        & 0.0800 & 0.1297  \\
UNet3D (Dice)           & 0.0598 & 0.0981  \\
\bottomrule
\end{tabular}
\label{tab:ablation}
\end{table}

\paragraph{Results} \Cref{tab:ablation} shows the results. We can draw the following conclusions. (1) Impact of Temporal Frame Interpolation (TFI): The top-performing method is UNet2D with a combination of Multi-Level Dice loss (ML-Dice), Input Cropping (IC), Output Cropping (OC), Augmentation (Aug), and Temporal Frame Interpolation (TFI), scoring highest in both mCSI (0.1069) and mF1 (0.1738). The inclusion of TFI in this configuration suggests its significant contribution to improving the model's performance.
(2) Effectiveness of Multi-Level Dice loss (ML-Dice): Models utilizing ML-Dice consistently outperform those using standard Dice loss. This indicates the effectiveness of ML-Dice in improving the model's accuracy. (3) Comparison of UNet2D and UNet3D Architectures: UNet2D models appear to outperform UNet3D models under similar configurations. This might suggest that convolution on the temporal dimension may not fully explore the temporal information.
(4) A clear trend shows that as more components (IC, OC, Aug) are added, the performance improves, regardless of the UNet version or loss function used. This underlines the importance of these preprocessing and augmentation techniques in enhancing model performance. Each additional component (IC, OC, Aug) contributes to a steady increase in both mCSI and mF1 scores. For example, comparing UNet3D (Dice) with each subsequent addition shows a gradual improvement, from 0.0598 to 0.0932 in mCSI and from 0.0981 to 0.1539 in mF1. (5) The least effective configuration is UNet3D with only Dice loss, while the most effective is UNet2D with the full suite of components including TFI. This wide range in performance across configurations underscores the impact of methodological choices on the model's predictive capabilities.

\begin{figure}[htp]
    \centering
        \begin{subfigure}[b]{0.49\linewidth}
            \includegraphics[width=\linewidth]{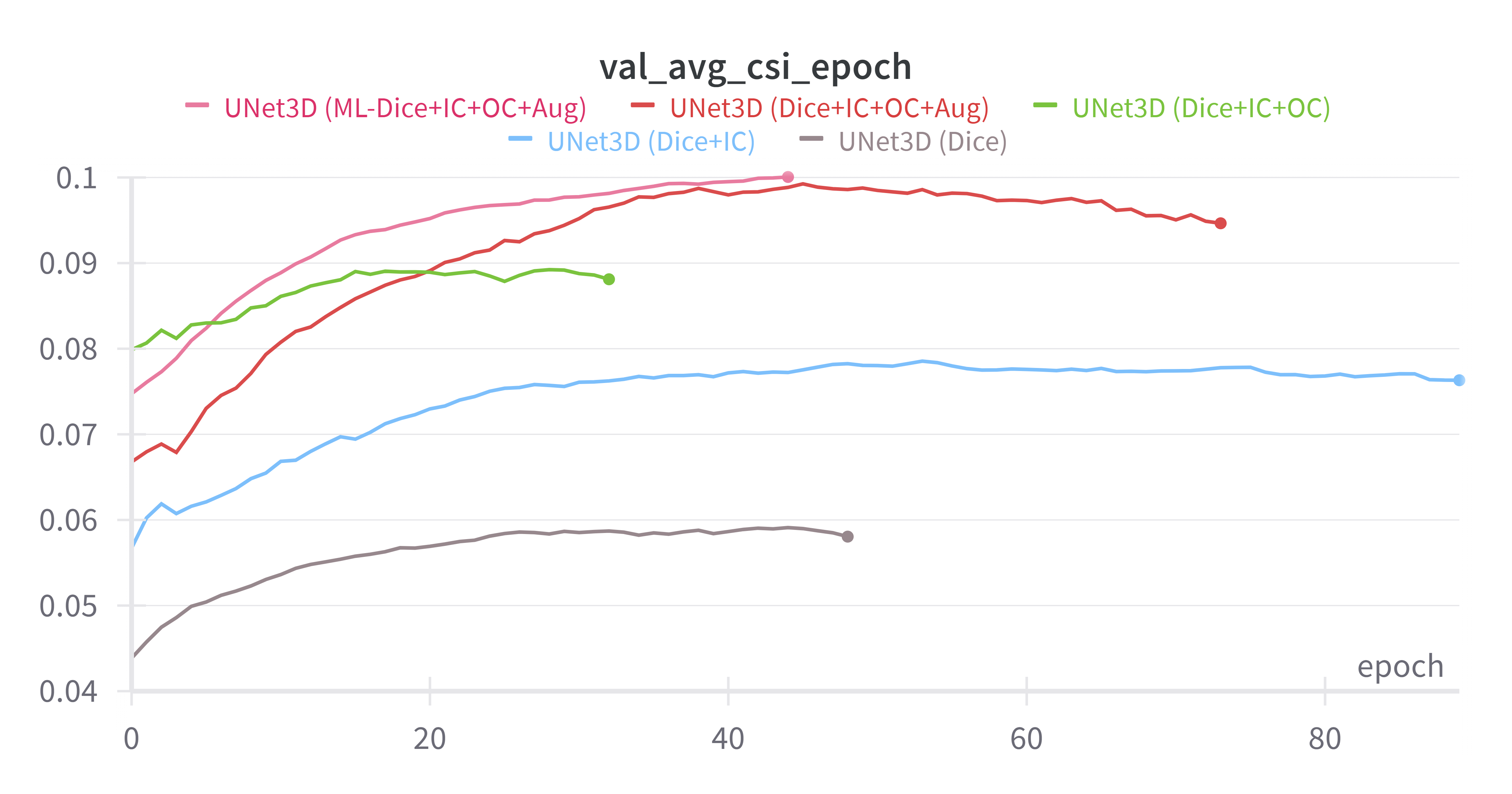}
    \caption{Ablation study on the UNet3D network.}
    \label{fig:u3d}
    \end{subfigure}
    \begin{subfigure}[b]{0.49\linewidth}
            \includegraphics[width=\linewidth]{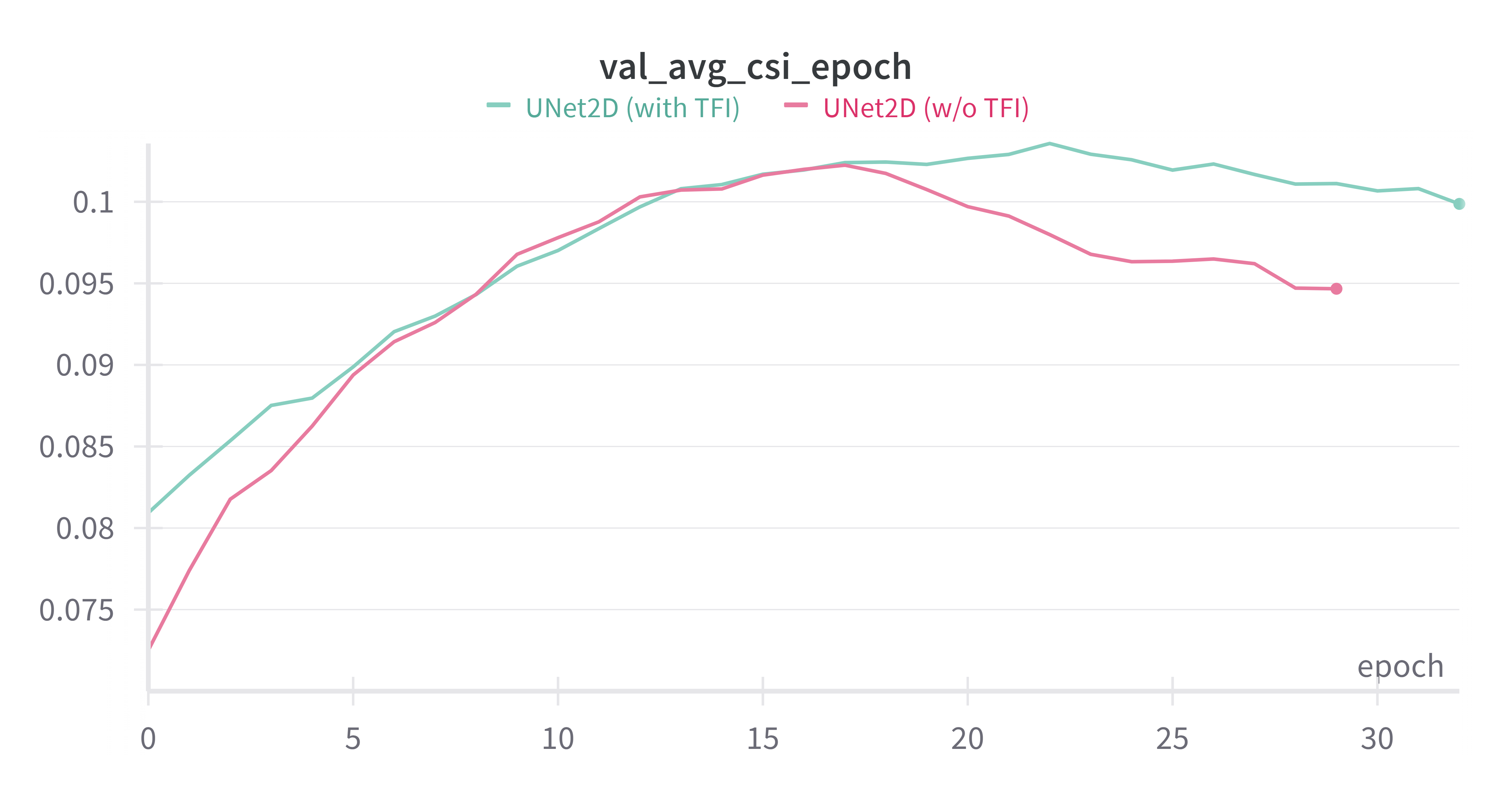}
    \caption{Ablation study on the UNet2D network.}
    \label{fig:u2d}
    \end{subfigure}
\end{figure}

\paragraph{Learning dynamics.} To show the influence of each component on the learning dynamics of the model. We plot the mCSI over epochs on two networks in~\cref{fig:u3d} and \cref{fig:u2d} respectively. The UNet3D model with the full suite of enhancements (ML-Dice+IC+OC+Aug) outperforms other configurations throughout the training process. It shows a steady increase in the validation mCSI over epochs. This indicates that the model is learning effectively from the training data when all enhancements are used. The configurations with intermediate numbers of enhancements (Dice+IC+OC+Aug and Dice+IC+OC) have similar trajectories, with the former consistently outperforming the latter, again highlighting the benefit of augmentation (Aug) in improving model performance. The configurations with fewer enhancements (e.g., Dice+IC+OC and Dice+IC) demonstrate lower performance, and the gap in performance widens as the training progresses. The model with only the Dice loss (UNet3D (Dice)) starts at the lowest performance and shows the least improvement over time, suggesting that the additional components contribute significantly to the model's learning capability. The presence of Temporal Frame Interpolation (TFI) in the UNet2D model shows a clear advantage over the model without TFI. The model with TFI maintains a higher average CSI throughout the training epochs.

In conclusion, the ablation results from the figures and the table demonstrate that the combination of TFI, ML-Dice and the selected training strategies significantly enhances the performance of UNet models for the task of rainfall prediction. The proposed TFI and ML-Dice also alleviate the overfitting on training data and accelerate the convergence of the model.
\section{Conclusions}
Over the past years, accurate predictions of rain events have become ever more critical, with climate change increasing the frequency of unexpected rainfall. 
\textit{Weather4cast} 2023 competition aims to build machine learning models with the ability of super-resolution rain movie prediction under spatio-temporal shifts. 
This year's metric requires finer-grain prediction of the precipitation, posing an even bigger challenge to the model's robustness.
In this paper, we propose a solution with a newly proposed temporal frame interpolation that enhances the robustness by creating synthetic between sampling intervals.
We also propose the multi-level dice loss that takes into account the ordinal relationship between rainfall rates, improving the prediction accuracy.
By combining several explored training strategies and proposed TFI and multi-level dice loss, our model can achieve the best result on the \textit{transfer learning leaderboard of Weather4cast'23 competition}, showing the practical applicability and superiority of our method over existing approaches under rigorous and competitive conditions. 
In conclusion, the Temporal Frame Interpolation method, supported by newly proposed ML-Dice loss and data augmentation strategy, offers a promising direction for future developments in high-dimensional weather forecasting tasks.
We hope that our contributions will serve as a stepping stone for subsequent innovations in the field.

{\small
\bibliographystyle{plain}
\bibliography{main}
}

\end{document}